# Exploiting the capacity of deep networks only at training stage for nonlinear black-box system identification

**Vahid MohammadZadeh Eivaghi**

Department of Electrical Engineering K. N. Toosi University of Technology Tehran, Iran, email: vmohammadzadeh@email.kntu.ac.ir

**Mahdi Aliyari Shoorehdeli**

Department of Electrical Engineering K. N. Toosi University of Technology Tehran, Iran, email: aliyari@kntu.ac.ir

**Abstract**

To benefit from the modeling capacity of deep models in system identification, without worrying about inference time, this study presents a novel training strategy that uses deep models only at the training stage. For this purpose two separate models with different structures and goals are employed. The first one is a deep generative model aiming at modeling the distribution of system output(s), called the teacher model, and the second one is a shallow basis function model, named the student model, fed by system input(s) to predict the system output(s). That means these isolated paths must reach the same ultimate target. As deep models show a great performance in modeling of highly nonlinear systems, aligning the representation space learned by these two models make the student model to inherit the approximation power of the teacher model. The proposed objective function consists of the objective of each student and teacher model adding up with a distance penalty between the learned latent representations. The simulation results on three nonlinear benchmarks show a comparative performance with examined deep architectures applied on the same benchmarks. Algorithmic transparency and structure efficiency are also achieved as byproducts.

## 1. Introduction

System identification is an active field in automatic control aiming at finding a dynamic mapping based on I/O data collected from a real-world system, for such purposes as analysis, control, simulation, prediction, and diagnosis [1], [2], [3], and [4]. Existing algorithms try to adjust the parameters of an adaptive model to meet some criteria that are defined based on the modeling purpose. Based on the type of relationship between the input(s) and output(s) of a system, identification algorithms are divided into linear and nonlinear methods. Although linear system identification methods can approximate a wide range of real-world systems, they will be imprecise in modeling systems whose time-varying and nonlinear behavior cannot be ignored [5]. As a result, nonlinear identification methods come into the picture. Mathematically speaking, linear methods try to find a linear combination of basis vectors defined as a lagged-window of the input(s) and output(s) of system to decode the output, meaning they are working in the subspace of model input(s). On the contrary, nonlinear methods try to find an intermediate space at which the system output can be easily constructed from a nonlinear mapping of shifted versions of input(s) and output(s) of systems. As a result, it can be said they are different in the way they find an intermediate space. This perspective on identification methods collide with machine learning view, as one of key challenges in machine learning is also to find an appropriate intermediate space, or rather representation, where the task is straightforward. This fact illuminates the massive potential of transferring the model architectures and training paradigms of machine learning into the system identification domain, which has been recently surveyed in [6]. Machine learning is an umbrella term for solving problems through learning from data [7]. It is currently applied to various applications like natural language processing (NLP), computer vision, and time series analysis.

Due to the increasing demand for high-performant intelligent systems, deep learning has emerged as a new discipline and, nowadays, has become a dominant approach for modelling highly complex systems. Deep learning algorithms build an enriched representation space from the original input space reflecting important explanatory factors sufficient for the task. One way to obtain such a representation is to use auto-encoders (AE) [8]. Auto-encoders consisting of an encoder layer and decoder layer, learn a deterministic map to encode the original input space into a lower dimensional space and convert it back [9]. Due to the mathematical similarity between auto-encoders and state space models, auto-encoders are mostly used in system identification domains. In [10], an AE-based model is proposed to identify an LTI state space model. Given the state space dimension, the adopted AE consist of a nonlinear encoder and a linear decoder. The nonlinear encoder takes a lagged-window of input(s) and output(s) and produce the state space and observation matrices, using which the linear decoder tries to estimate the system output. In [11], an auto-encoder based model approximates a nonlinear state space model. They adopt a special topology to provide acceptable performance for both open-loop and closed-loop applications. Since it is assumed there is no information regarding the model order, an additional $L$1-regularization term is applied to the first layer of the encoder and decoder in the hope that unimportant dynamics will be automatically removed. However, it will not work since the input variables are correlated in time, yet LASSO is applicable for independent variables [12].

Another approach for learning representation is latent variable models [8]. Latent variable models are a specific class of generative models. They assume that an underlying mechanism explains the variation behind the data. Energy-based models are a specific class of latent variables that approximate the joint distribution over some variables. They are undirected graphical models decomposing a joint distribution as a product of some terms called potential functions parameterized using an exponential form of energy function. MLE solution for parameter estimation of the NARX model is often founded on the assumption that the distribution of output(s) given input(s) follow a Gaussian distribution which may be violated in practice. In [13], an energy-based NARX model is proposed for nonlinear system identification. They provide a general formulation for NARX models by employing energy-based models whose energy function is parameterized using a deep neural network. Although the developed model can model the complex conditional distribution, the computational burden required for approximating the partition function of energy-based models is heavy. Variational auto-encoder (VAE) is another class of latent variable models which are also used for nonlinear system identification. In [14], a structured form of prior is introduced to incorporate the physical information of system into VAE training for identifying systems formed by PDE equation. In [15] some variants of dynamical VAE models are used to identify nonlinear state space models. Dynamical VAE is a class of sequential latent variable models, surveyed in [16], that is formed by combining a dynamical model like RNN and VAE. In [17], an AE-based approach is proposed to identify the nonlinear state space models. Despite the previous works in this direction, [17] only provides some valuable theorems to support their idea. Like the linear subspace method, they divide the input and output space into past and future. The past input and output are fed into an encoder to estimate the current state; then using the current state and input future, the decoder is responsible for decoding the output future. They also show that in the case of linear relationships, the proposed methodology will result in the same formula as subspace method.

Another direction of utilizing deep learning in system identification is using temporal convolutional networks (TCN) to identify nonlinear I/O relationships [18]. It is shown that, from mathematical perspective, TCNs are the same as block-oriented models. In [19] and [20], deep NARX and NFIR are examined. [21] proposes a model so-called deep Lagrangian network incorporating physics-related information into the training process of a deep network. The utilized approach aims to find the state space representation of mechanical systems, having known order and measurable states, through Euler–Lagrange equations approximated using two different deep neural networks, one is responsible for approximating the mass matrix and another for estimating the potential energy. Accordingly, the parameters of the model are adjusted to minimize the residual of the Euler-Lagrange equation. Similar work is also extended for approximating the Hamilton equation in [22] using deep neural network that is called deep Hamiltonian network.

Although using deep models for nonlinear system identification shows a remarkable performance, they will dictate limitation in applications where processing speed and memory usage are essential. In other words, based on reported works, deep models give a comparative performance at a lower speed and higher memory footprint than their traditional counterparts. It is possible to employ some compression methods for optimizing resource usage by deep models [23] and [24], yet it will increase the efforts required for modelling. This is not intended to criticize the use of deep learning in system identification domains. Still, it is going to illuminate some important aspects of their usage, which are not addressed in above-mentioned research. This important issue is the main motivation of the present work introducing a novel training approach that employs two different models: a deep generative networks, named the teacher model, and a shallow MLP model, named the student model. The teacher model learn how to encode the marginal PDF of the output(s) of system, while the student model tries to model the conditional mapping of output(s) given the input(s) of the system. By relating these two different paths of output encoding at training time, knowledge will be transferred from the deep model to the shallow one, increasing its approximation capacity. The general view of the adopted approach is given in Figure 1.

To the best of the authors' knowledge, this is the first work that uses a deep model to guide a shallow predictor to produce an appropriate representation in the context of nonlinear system identification. The proposed approach pave the way for harnessing the effectiveness of deep networks for making a shallow model without increasing the inference time and computational burden, as our contribution. Furthermore, it can be seen as a gentle step toward the interpretability of the nonlinear black-box model. The adopted approach make the model interpretable at the level of the training algorithm [25], since the input mapping obtained by basis function models contains summary statistics of the system output(s), informing us of what model learns during the training phase.

The rest of paper is organized as the following. In the next section, concepts related to basis function model and generative models are shortly mentioned. In section 3, the problem formulation is presented. In section 4, the effectiveness of the proposed approach is examined on three nonlinear benchmarks. Finally, in section 5, some conclusions are drawn.

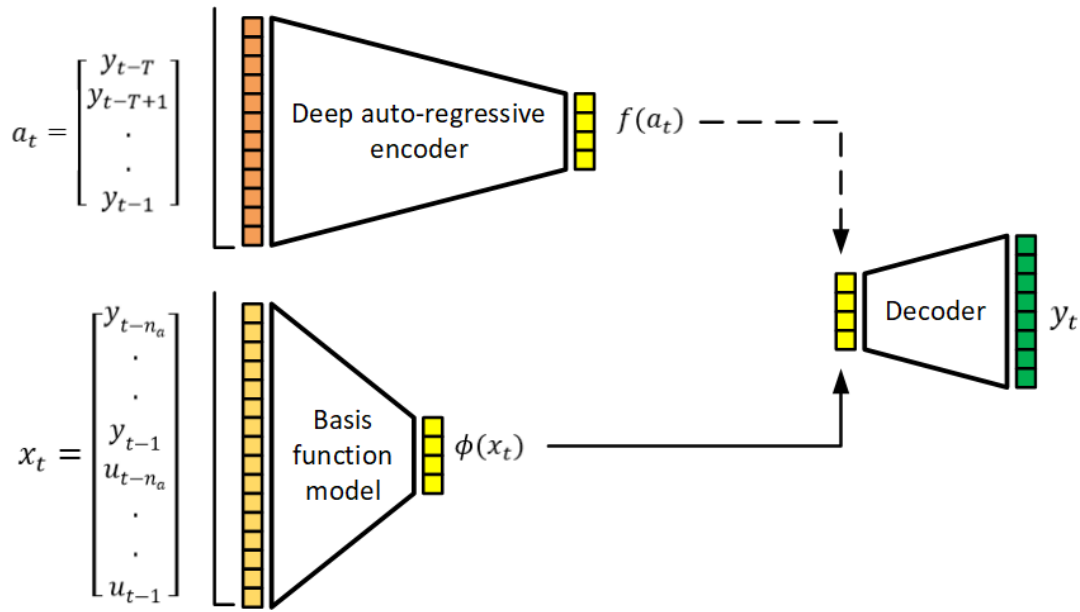

Figure 1 – The general view of proposed approach – the pair deep teacher-shallow student is simultaneously trained in a way that representations $\phi(x_t)$ and $f(a_t)$ are maximally aligned and the predictability of $y_t$ from both $\phi(x_t)$ and $f(a_t)$ increase. Upon completing the training process, the deep generative model will cease generating dashed lines

## 2. Background tools

In this section, a general definition of the presented work is given. Initially, the basis function formulation will be reviewed, and then the mathematical background of various technical tools and technologies utilized in this study will be introduced.

### 2.1. Basis function models

Given the data $D = \{x_t, y_t\}_{t=1}^{T}$ collected from a real-world system, the system identification problem is to model $P_\theta(y_t|x_t)$ which $x_t$ is a window of shifted input(s), output(s), and prediction error(s). For nonlinear models, the output(s) $y_t$ relate to the input(s) $x_t$ using a nonlinear function driven by an additive noise:

$$y_t = f_\theta(x_t) + e_t, \quad E(e_t) = 0, E(e_t e_t^T) = R \tag{1}$$

So the conditional distribution $P_\theta(y_t|x_t)$ is a Gaussian distribution whose mean and covariance are $f_\theta(x_t)$ and $R$ respectively. From all realization of $f_\theta(x_t)$, almost all alternatives can be written in the following general form, called basis function models (BFM) formulation [1]:

$$y_t = \sum_{k=1}^{M} \alpha_k \phi_k(x_t; \theta) + e_t \tag{2}$$

Based on BFMs formulation, the output $y_t$ is modeled as a weighted average over M basis functions which must be nonlinear to realize a nonlinear model. There is no limitation on how the basis functions are modelled. They can be modelled by either black box models like a multi-layered perceptron, Gaussian process, polynomial approximation, functional expansion, and deep learning or transparent models like lookup tables, locally linear models, and fuzzy approximation. Black box models first map the input $x_t$ into a hidden space $\phi(x_t)$ at which the output prediction will be easily done. It is called hidden, since there is no clear observation on what model learns at this stage. Using the advancement of deep learning technologies like contrastive learning, however, one can guide the obtained hidden representation of input toward containing some useful information, easing the interpretation of learned representation.

*2.2. Deep generative models*

Generative modelling is a field in machine learning that tries to model the joint distribution over some variables either directly or indirectly. They can be used for two general purposes: improving the performance discriminative models and content generation. The former is the use-case of the generative model in this work. In contrast to discriminative models trying to find a mapping function from input to output using the discrepancies between existing patterns, generative models learn the statistical properties of each pattern [9]. The development of generative model is not limited to recent years and dates back to 1950 [26], starting with the development of Gaussian Mixture Model (GMM) and Hidden Markov Model (HMM) for modelling sequential data. However, their limited performance in modelling high-dimensional space and generating sequences with short-term dependency was frustrating. It was not until the advent of deep learning, stemming from proposing a fast greedy-strategy for training deep belief networks in 2006 [27], which generative models witness a great progress in performance, and deep generative models were born.

Generative models can be developed either through maximizing the likelihood function or using adversarial training, yet most existing models are developed based on the maximum likelihood approach. Adversarial training is a learning paradigm mostly used to create a robust machine learning model. Generative adversarial net and its variants [28] and [29] are well-known generative models developed based on adversarial training. Another direction for grouping generative models is to categorize them as latent variable or non-latent variables [30]. Latent variable models assume there is an underlying mechanism explaining the variation behind data. They are mostly used to model the data distribution, whereas non-latent variable models are only used to mimic sampling mechanism. Probabilistic PCA [12], Energy-based models [31], Variational Auto-encoder [32], diffusion models [33], and normalizing flow [34] are some examples of the latent variable models.

To develop a controlled basis function model, we must use models which give us the explicit notion of latent space. In other words, we need to have the learned representation by two different models in hand to make them maximally aligned. Although, any arbitrary instance of latent variable family can be selected, Variational RNN, dynamical version of VAE, is employed in this work. As the VAE and its dynamical version are previously used in the context of nonlinear system identification, for the sake of brevity we don't mention their mathematical details and refer the interested readers to [15], [16].

## 3. Proposed method

As it has been said, to harness the deep learning approximation power while remaining shallow at inference time, a pair of teacher-student network is adopted which the teacher model is a deep generative model and the student model is a shallow black-box network. These two models, at first, create two isolated paths for encoding the target, yet they link by making an alignment between the representations they learn. The proposed architecture is depicted in Figure 2.

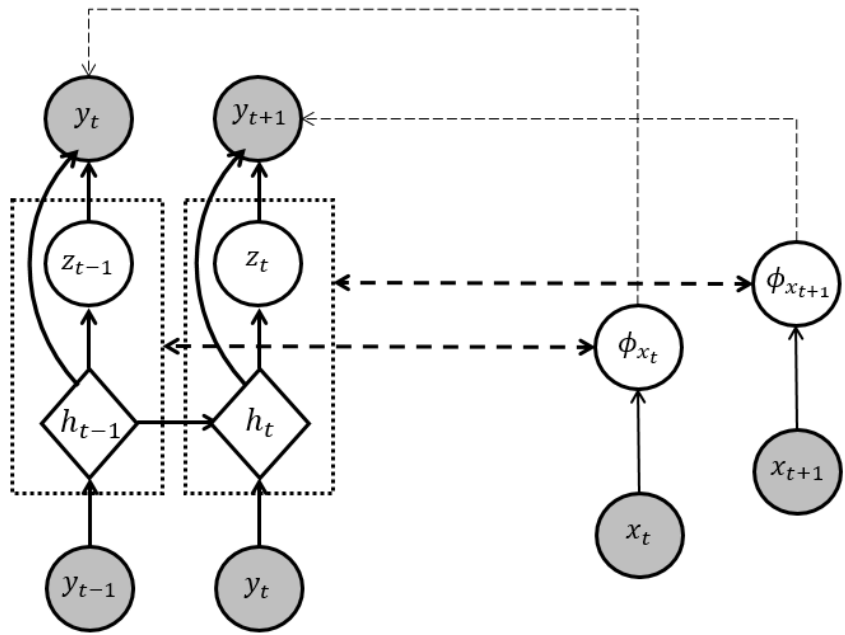

Figure 2 – Two-time slice Bayesian network of proposed scheme – the solid arrow line indicate direct dependency, the dashed double arrow line highlight the alignment between $\phi(x_t)$ and $[h_t; z_t]$ at each time step, and the dashed arrows show a separated direction of mapping of either $\phi(x_t)$ or $[h_t, z_t]$ to $y_t$ without interfering from either $[h_t, z_t]$ or $\phi(x_t)$ respectively. The left part of the architecture is the teacher model creating the representation $\phi_T(t) = [h_t, z_t]$ and the right part is the student model with learned representation $\phi_S(t) = \phi_{x_t}$.

As shown in Figure 2, the training scheme include three main parts. The left side of architecture is the teacher model, which is simply a VRNN model, and the right side shows the student model which is an MLP network. Two parts are coupled with each other through a dashed double arrow aligning the knowledge learned by the teacher model and the student model. In order for the architecture to be trained, we form a cost function based on the summation of three losses associated with each shown part.

*3.1. Teacher model – Dynamical VAE*

The generative path for encoding the distribution over the sequence $y_{1:T}$ is modeled using VRNN, a dynamical extension of the VAE model to sequential data. The original version of VAE is proposed for generating static data like images. With increasing demand for modelling the dynamic world, some works have commenced to generalize VAE to sequential data. Deep state space models, known as Deep Kalman Filter as well, are of the earlier models for this purpose [16]. Deep state space model is simply a nonlinear state space models whose state transition and observation models are parameterized using deep neural networks [16]. This development is followed by introducing such architectures as VAE-RNN [16], Variational RNN (VRNN) [35], SToRN (Stochastic Recurrent Network) [36], and KVAE (Kalman Variational Auto-encoders) [37], Stochastic RNN (SRNN) [38], Recurrent VAE (RVAE) [39], Disentangled Sequential Auto-encoder (DSAE) [40], all are surveyed in [16], as dynamical VAE. Unlike "static" VAE and similar to SSM, observations and latent variables should be considered to be temporally correlated and can have more or less complex dependencies across time. A general framework for modeling and training dynamical VAE is first given in [16] from which the aforementioned structures can be instantiated.

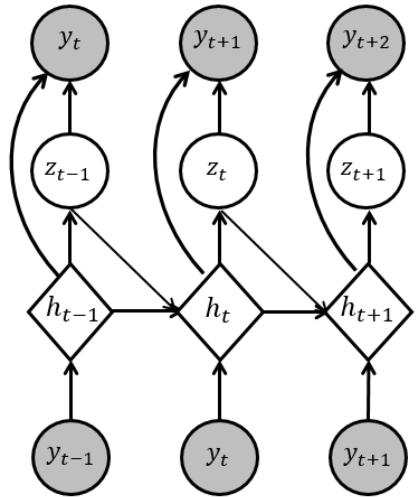

Figure 3 – Variation Recurrent Neural Network (VRNN) – the shadow circles indicate the observation, $h_t$ is deterministic temporal information and $z_t$ is the state of VAE.

VRNN model first maps the input sequence into a hidden space using a variant of the RNN model, in our case GRU, then passes the obtained hidden space into a VAE. The concatenation of VAE and RNN states is used for producing output sequences. The graphical representation of VRNN is depicted in Figure 3. The probability distribution over the variables is factorized as the following:

$$P_{\theta_T,\theta_C}(y_{1:T}, h_{1:T}, z_{1:T}|h_0)$$
$$= \prod_{t=1}^{T} P_{\theta_C}(y_t|h_t, z_t) P_{\theta_T}(z_t|h_t) P_{\theta_T}(h_t|h_{t-1}, y_{t-1}, z_{t-1}) \quad (3)$$

Where $h_0$ is the initial state of the RNN model and can be set to a zero vector. Since the update equation for the hidden state of RNN model is a deterministic function, $P_{\theta_T}(h_t|h_{t-1}, y_{t-1}, z_{t-1})$ is a degenerative distribution. Other conditional terms are also modelled as a Gaussian distribution whose mean and covariance matrices are calculated using a deep neural network as the following:

$$P_{\theta_T}(h_t|h_{t-1}, y_{t-1}, z_{t-1}) = \delta(h_t - \tilde{h}_t), \quad \tilde{h}_t = f^h_{\theta_G}(y_{t-1}, h_{t-1}, z_{t-1}) \quad (4)$$

$$P_{\theta_T}(z_t|h_t) \sim N\left(\mu^z_{\theta_G}(\tilde{h}_t), \Sigma^z_{\theta_T}(\tilde{h}_t)\right), \begin{cases} \mu^z_{\theta_T}(\tilde{h}_t) = NN^z_{\theta_T}(\tilde{h}_t) \\ \log \Sigma^z_{\theta_T}(\tilde{h}_t) = NN^z_{\theta_T}(\tilde{h}_t) \end{cases} \quad (5)$$

$$P_{\theta_C}(y_t|\phi_T(t) = [h_t, z_t]) \sim N\left(\mu^{yzh}_{\theta_C}(\phi_T(t)), \Sigma^{yzh}_{\theta_C}(\phi_T(t))\right), \begin{cases} \mu^{yzh}_{\theta_C}(\phi_T(t)) = NN^{yzh}_{\theta_C}(\phi_T(t)) \\ \log \Sigma^{yzh}_{\theta_C}(\phi_T(t)) = NN^{yzh}_{\theta_C}(\phi_T) \end{cases}$$

(6)

*3.2. Student model – black-box basis function model*

The student model is considered a time-delayed neural network (TDNN) fed by the lagged version of input(s) and output(s) of systems. There is nothing specific about using TDNN except that the output layer of which is shared with VRNN decoder. If we consider the hidden representation of the student model is denoted as $\phi_S(x_t)$ where $x_t = [u_{t-n_b}, \ldots u_{t-1}, y_{t-n_a}, \ldots y_{t-1}]$ the parameterization of the output layer will be the same as the VRNN decoder.

$$P_{\theta_C}(y_t|\phi_S(x_t)) \sim N\left(\mu^{ys}_{\theta_C}(\phi_S(x_t)), \Sigma^{ys}_{\theta_C}(\phi_S(x_t))\right), \begin{cases} \mu^{yzh}_{\theta_C}(\phi_S(x_t)) = NN^{yzh}_{\theta_C}(\phi_S(x_t)) \\ \log \Sigma^{yzh}_{\theta_C}(\phi_S(x_t)) = NN^{yzh}_{\theta_C}(\phi_S(x_t)) \end{cases}$$

(7)

*3.3. Alignment link*

If it is assumed that the experiment designed for system identification covers all possible values of output(s) of systems, one can expect that the generated output(s) by the student model belong to the distribution encoded by teacher model. To achieve this, we employ a Variational RNN (any arbitrary generative model with explicit latent space can be used), as teacher model, to model the distribution of output(s), and a simple MLP network, as the student model, to map input(s) to output(s). Both networks can be decomposed into two parts, one for creating a representation and another for a projection head to create the output prediction. If we share the head projection between two networks, the distribution over the output encoded by two networks given the learned representation will:

$$p_{\theta_C}[y_T|\phi_T] \sim N\left(\mu_{\theta_C}(\phi_T), \Sigma_{\theta_C}(\phi_T)\right), \qquad p_{\theta_C}[y_S|\phi_S] \sim N(\mu_{\theta_C}(\phi_S), \Sigma_{\theta_C}(\phi_S))$$

Where $y_i$ and $\phi_i$ are the output and representation created by model $i$ for $i = T(teacher), S(student)$, and $\theta_C$ is the shared parameter between student and teacher model. In order for generated samples $y_S$ to belong to the distribution $p_{\theta_C}[y_T|\phi_T]$, all possible modes and variations in $p_{\theta_C}[y_S|\phi_S]$ must be covered by $p_{\theta_C}[y_T|\phi_T]$, meaning the KL-divergence $D_{KL}\left(p_{\theta_C}[y_S|\phi_S] \| p_{\theta_C}[y_T|\phi_T]\right)$ should approach zero. By definition we have:

$$\begin{aligned} D_{KL}&\left(p_{\theta_C}[y_S|\phi_S] \| p_{\theta_C}[y_T|\phi_T]\right) \\ &= \frac{1}{2}\left\{\log \frac{\Sigma_{\theta_C}(\phi_T)}{\Sigma_{\theta_C}(\phi_S)} - D + \left(\mu_{\theta_C}(\phi_T) - \mu_{\theta_C}(\phi_S)\right) \Sigma_{\theta_C}^{-1}(\phi_T) \left(\mu_{\theta_C}(\phi_T) - \mu_{\theta_C}(\phi_S)\right)^T \right. \\ &\quad \left. + trace\left(\Sigma_{\theta_C}^{-1}(\phi_T)\Sigma_{\theta_C}(\phi_S)\right)\right\} \end{aligned}$$

Where D is the dimension of the covariance matrix. Since the head projection is shared between student and teacher model, the parameterization of mean and covariance for both model differs only in their inputs. Therefore, the KL-divergence term will approach zero if we train both models simultaneously in a way that results in $\phi_T = \phi_S$. This motivated us to use the contrastive loss to make a bridge between two models using a distance penalty $|\phi_T - \phi_S|_2^2$, meaning the representation knowledge learned by the teacher model will be inherited by the student network provided that the following cost function approach to zero:

$$\begin{aligned} J(\theta_T, \theta_S, \theta_C) = \frac{1}{T}\Bigg\{ &\sum_{t=1}^{T} L\left(y_t; f_{\theta_C}(\phi_T; \theta_C)\right) + L_T + L\left(y_t; f_{\theta_S}(\phi_S; \theta_C)\right) \\ &+ |\phi_T(y_{t-\tau:t-1}; \theta_T) - \phi_S(y_{t-n_a:t-1}, u_{t-n_b:t-1}; \theta_S)|_2^2 \Bigg\} \end{aligned} \qquad (8)$$

Where $L$ is the cost of the projection head and $L_T$ is additional terms associated with the teacher model. An Alternative solution can also be considered by replacing the distance penalty with the correlation term $-\phi_T^T \phi_S$.

The probabilistic view of the cost function (8) will be as (9), where the first term is the cost of student network, the second terms is the cost of teacher network, and the third one is related to the alignment link, optimized using gradient decent algorithm.

$$L_M(\theta_T, \theta_S, \theta_C, \lambda)$$

$$= -\alpha_1 E_\phi \left( \sum_{t=1}^{T} \log P_{\theta_C, \theta_S}(y_t | \phi_S) \right)$$

$$- \alpha_2 E_{q_\lambda} \left( \sum_{t=1}^{T} \log P_{\theta_C}(y_t | \phi_T) + D_{KL}\left(q_\lambda(z_t | y_t, \widetilde{h}_t) \big| P_{\theta_T}(z_t | \widetilde{h}_t)\right) \right) \quad (9)$$

$$+ \alpha_3 |\phi_T - \phi_S|_2^2$$

$$\theta_T^*, \theta_S^*, \theta_C^* \lambda^* = \arg \min_{\theta_T, \theta_S, \theta_C, \lambda} L_M(\theta_T, \theta_S, \theta_C, \lambda) \quad (10)$$

## 4. Numerical experiments

In this section, the proposed approach is evaluated on three experiments. In order for us to conduct a comparative study, for each experiment two different models are trained, a baseline model denoted by $M_B$ consisting of a single MLP model, and a regenerative model indicated by $M_R$ including a pair of teacher-student networks. $M_B$ is trained ordinarily without intervention from a generative teacher while $M_R$ is trained using the proposed approach. The hyper-parameters of each model are indicate by $H$ followed by the associated subscript. What is more, they are represented as a tuple whose length is the depth of the network, and each element of which is the width of the network. The type of layers is fully connected layer except for the first layer of teacher network, which is GRU for recurrence computing.

For training stage, hyper parameters tuning and model selection, the data is split in training and validation set. An ELBO approach is adopted to select optimal values of depth and width of the networks. We also constrain $H_S$ to be a subset of $H_B$ ($H_S \subseteq H_B$) meaning the number of parameters of the student model in the worst case are as large as the baseline model. A separate test data set is also used for calculating the final performance. The ADAM optimizer with default parameters and a constant learning rate 0.001 is used. Early stopping is also applied to monitor the evaluation metrics.

Three experiments conducted in this works are exactly the same as experiments done in [15]: 1) linear Gaussian model, 2) Narendra-Li benchmark [41], and 3) Wiener-Hammerstein process noise [42]. Root mean square error (RMSE) is considered as performance metric, $\sqrt{\frac{1}{N_T}\sum_{t=1}^{N_T}(y_t - \hat{y}_t)^2}$ where $N_T$ is the total number of test samples and $\hat{y}_t$ is the model output. Furthermore, to quantify the quality of uncertainty estimate we use negative log-likelihood as $-\frac{1}{N_T}\sum_{t=1}^{N_T} \log N(y_t | \mu_t^{de}, \Sigma_t^{de})$ indicating how likely a given true sample point $y_t$ is occurred under the model distribution.

### 4.1. Linear Gaussian model

Consider the following linear state space model [15] with process noise $w_k \sim N(0, 0.5I)$ and measurement noise $v_k \sim N(0, I)$.

$$x_{k+1} = \begin{bmatrix} 0.7 & 0.8 \\ 0 & 0.1 \end{bmatrix} x_k + \begin{bmatrix} -1 \\ 0.1 \end{bmatrix} u_k + w_k \qquad (11)$$

$$y_k = \begin{bmatrix} 1 & 0 \end{bmatrix} x_k + v_k \qquad (12)$$

The model is trained and validated with 50000 samples collected from the given system and excited by random uniform signal in the interval [-2.5, 2.5] to mimic the I/O behavior of the system. The sequence length of input and output for forming the input of model is set to 10 and 5 respectively, meaning:

$$x_t = [u_{t-10}, \dots, u_{t-1}, y_{t-5}, \dots, y_{t-1}]$$

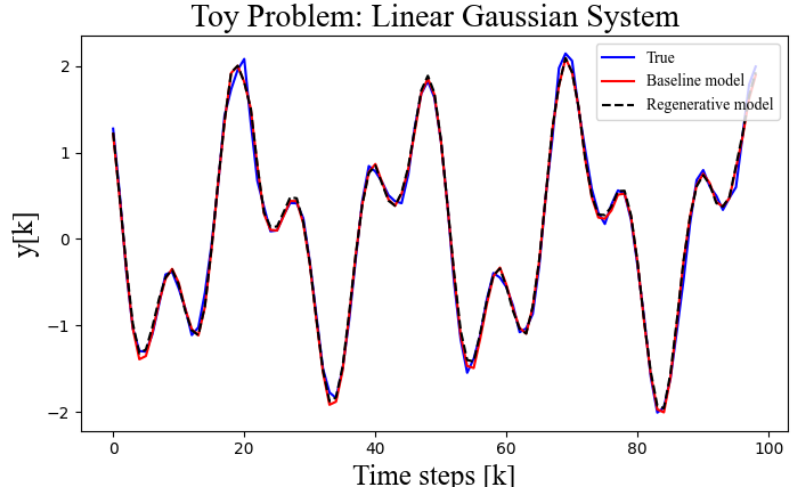

Figure 4 – Toy problem: Linear Gaussian systems – result for prediction performance of the identified model on test data

A grid search is applied for hyper-parameter tuning of the baseline model with depth = {1, 2, 3, 4, 5} and width = {10, 20, 30, 40, 50, 60, 70, 80, 90, 100}. Here, the configuration (60, 30), two layers with 60 and 30 neurons, is selected as the optimal hyper-parameters for the baseline model, meaning the shape of the baseline network would be $(15, 60, 30, 1)$. In addition, we choose $(15, 30, 1)$ and $(1, 15, 60, 30, 1)$ as the configurations of the student model and the teacher network sharing the last two layers (30, 1). Since the teacher network will be skipped at inference stage, it is clear that the student model has 1800 fewer parameters than the baseline model. However, this is achieved in exchange for increasing the training cost.

Considering the test signal as $u_k = \sin\frac{2k\pi}{10} + \sin\frac{2k\pi}{5}$, the output of the final model obtained by averaging over 100 identified models for both baseline and regenerative models is depicted in Figure 4. The value of RMSE is calculated as 0.068 and 0.077 for the baseline model and the regenerative model respectively, illuminating the effectiveness of proposed approach. It is important to mention that different configurations yield the same level of performance, which is attributed to the generative teacher guiding the representation of the basis function model to contain predictive information for decoding the output.

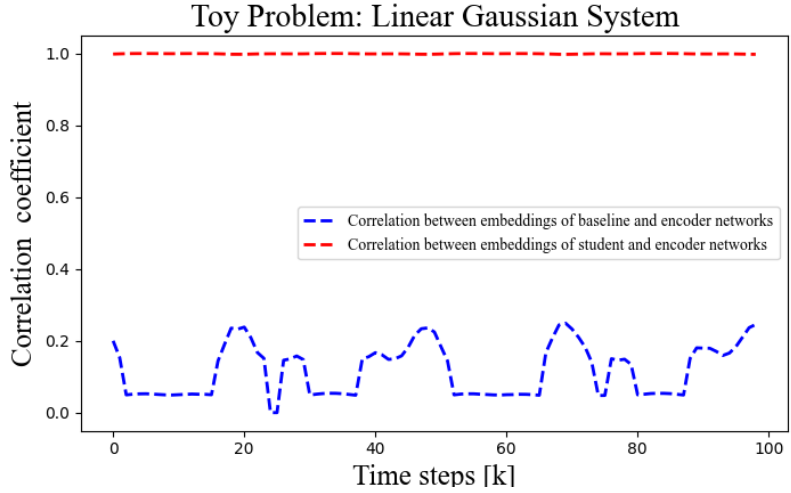

Figure 5 – Toy problem: Linear Gaussian systems – correlation coefficient between obtained representations obtained by baseline and regenerative model

Having an insight into the inner layers of both the baseline and regenerative models, it is clear that the representation obtained by the student model, acting on $x_t$, is highly correlated with the generative representation, obtained by teacher network from the lagged version of $y_t$, while this is not the case for the baseline model, Figure 5. That means the output can be easily approximated using a linear decoder shared between the teacher and student model, forcing the student model to approximate the deep encoder using a shallow architecture.

*4.2. Narendra-Li benchmark*

Narendra-Li is designed as a highly nonlinear benchmark but fictional system. Its dynamic equations are described as the following:

$$\begin{bmatrix} x_{k+1}^1 \\ x_{k+1}^2 \end{bmatrix} = \begin{bmatrix} \left(\frac{x_k^1}{1+(x_k^1)^2}\right)\sin x_k^2 \\ x_k^2 \cos x_k^2 + x_k^1 \exp\left(-\frac{(x_k^1)^2+(x_k^2)^2}{8}\right) + \frac{u_k^3}{1+u_k^2+0.5\cos(x_k^1+x_k^2)} \end{bmatrix} \quad (13)$$

$$y_k = \frac{x_k^1}{1+0.5\sin x_k^2} + \frac{x_k^2}{1+0.5\sin x_k^1} + e_k \quad (14)$$

Where $e_k$ is measurement noise $e_k \sim N(0, I)$.

The model is trained and validated with the same settings applied for identifying linear Gaussian model, and the same procedure is considered for tuning the model hyper-parameters. The sequence length of input and output for forming the model input is set to 20 and 5 respectively. The optimal shape of baseline networks is selected as (25, 45, 45, 10, 1). The configurations (25, 45, 10, 1) and (1, 25, 45, 45, 10, 1) are respectively considered for the student and teacher model sharing the last two layers (30, 1).

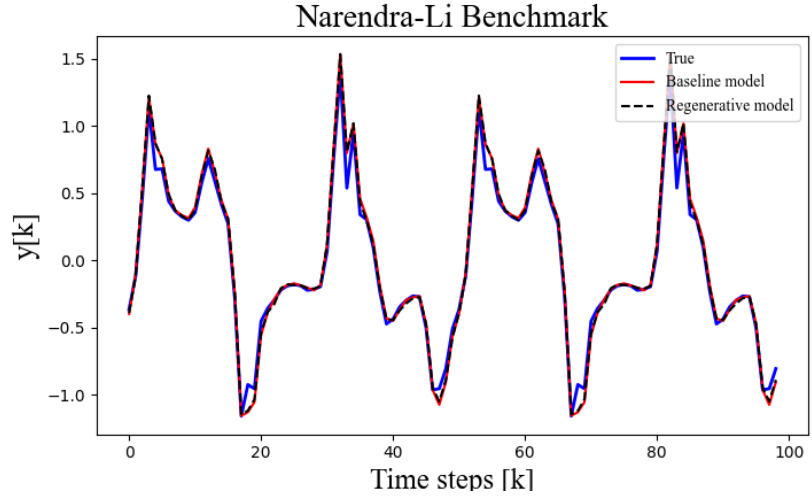

Figure 6 – Narendra-Li Benchmark – result for prediction performance of the identified model on test data

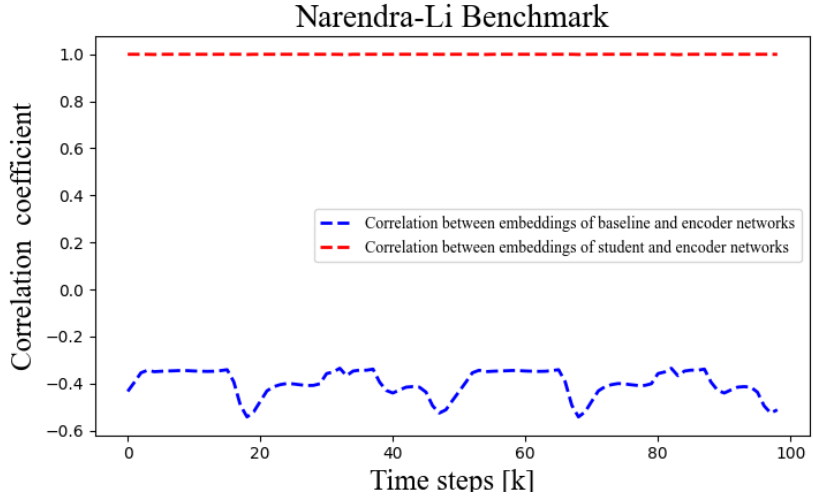

Figure 7 – Narendra-Li benchmark – correlation coefficient between obtained representations obtained by baseline and regenerative model

The output of identified model is illustrated in Figure 6. The value of RMSE is calculated as 0.091 and 0.083 for the baseline and regenerative model respectively. The correlation plots between the obtained representations and the encoder network are also shown in Figure 7.

*4.3. Wiener-Hammerstein with process noise*

Wiener-Hammerstein (WH) process is simulated using an electric circuit by [42] where the process noise enters before applying nonlinearity modelled by a diode-resistor network. The training and validation data are partitioned based on available 64162 samples where the input is swept sine. The sequence length of input and output for forming the model input both are set to 20, meaning:

$$x_t = [u_{t-20}, \ldots, u_{t-1}, y_{t-20}, \ldots, y_{t-1}]$$

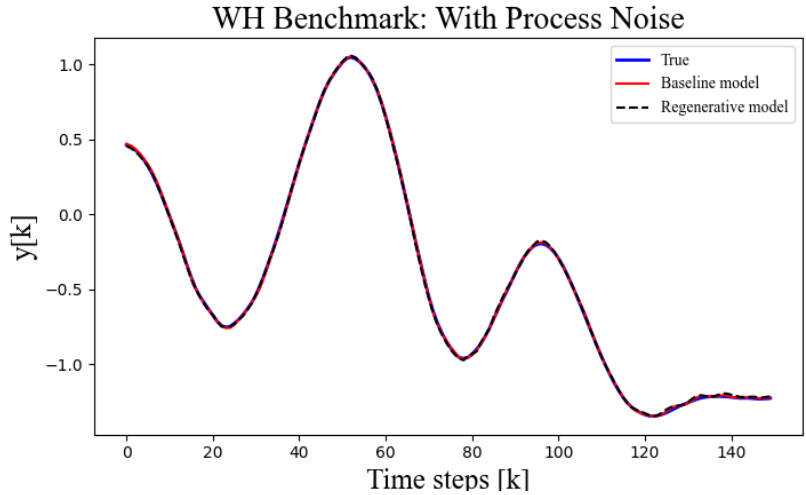

Figure 8 – WH benchmark with process noise – result for prediction performance of the identified model on test data (faded multi-sine signal)

Similar to the two previous simulations, applying the grid search, the configuration shape of the baseline model is regarded as $(40, 80, 20, 1)$. Likewise, the shapes $(40, 20, 1)$ and $(1, 40, 80, 20, 1)$ are respectively considered for student and student network sharing the last two layers $(20, 1)$.

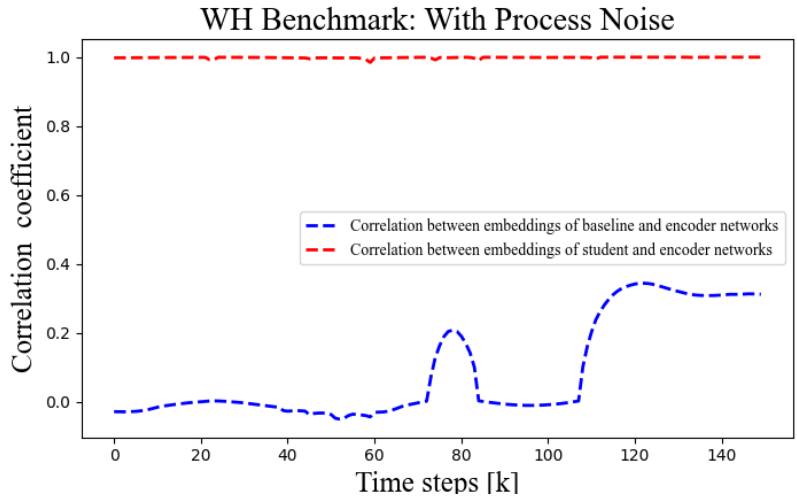

Figure 9 – WH benchmark with process noise – correlation coefficient between obtained representations obtained by baseline and regenerative model

The model is evaluated on 8192 samples collected by exciting the system using faded multi-sine, depicted in Figure 8. The value of RMSE is calculated as 0.02 and 0.07 for the baseline and student model respectively. The correlation plots between the obtained representations and the encoder network are also shown in Figure 9. The obtained results are also summarized in Table 1.

Table 1 – Results for regenerative model and baseline model

| Model \ EXP | Baseline model | | | Regenerative model | | |
|---|---|---|---|---|---|---|
| | RMSE | NLL | Architecture | RMSE | NLL | Architecture |
| Toy LGSSM | 0.068 | – | (15, 60, 30, 1) | 0.077 | 0.94 | (15, 30, 1) |
| Narendra-Li | 0.091 | – | (25, 45, 45, 10, 1) | 0.083 | 1.01 | (25, 45, 10, 1) |
| WH | 0.02 | – | (40, 80, 20 1) | 0.09 | 0.991 | (40, 20 1) |

## 5. Conclusion

Although utilizing deep networks for system identification extends existing tools and improves the identification performance, their over-parameterized nature is a limiting factor for real-time applications. This paper presents a training approach that leverages the modelling capacity of deep networks exclusively during the training phase. The adopted approach uses a pair of teacher-student networks where the teacher network is a generative model encoding the probability distribution over the output sequences of system, and the student model is a simple black-box basis function model. Both models are concurrently trained to ensure that the representation of the student model is maximally aligned with the generative model. Not only does this procedure make the conditional mapping from input to output easier, the representation learned by basis function model provide uncertainty quantification and algorithmic transparency as well, since the learned representation by student model contains statistics of system output(s), in spite of the traditional black-box models which are not clear in what they are learning. Simulation results on three different experiments, summarized in Table 1, show that the adopted approach will bring model compression as byproduct, as the student network has fewer number of parameters than a basis function model that is trained ordinarily to give us the same level of performance.

## 6 – References